\documentclass[lettersize,journal]{IEEEtran}
\usepackage{amsmath,amsfonts}
\usepackage{algorithmic}
\usepackage{array}
\usepackage[caption=false,font=normalsize,labelfont=sf,textfont=sf]{subfig}
\usepackage{textcomp}
\usepackage{stfloats}
\usepackage{url}
\usepackage{verbatim}
\usepackage{graphicx}
\usepackage{orcidlink}
\def\BibTeX{{\rm B\kern-.05em{\sc i\kern-.025em b}\kern-.08em
    T\kern-.1667em\lower.7ex\hbox{E}\kern-.125emX}}
\usepackage{balance}

\usepackage{algorithm}
\usepackage{algorithmic}
\usepackage{amsmath}

\begin{document}
\title{Reconsidering Overthinking: Penalizing Internal and External Redundancy in CoT Reasoning}
\author{
    Taihang Zhen \orcidlink{0009-0008-2802-4043}, 
    Jialiang Hong,
    Kai Chen \orcidlink{0009-0004-8764-5131},
    Guang Yang \orcidlink{0009-0005-4937-0568},
    Junlan Feng \orcidlink{0000-0001-5292-2945}, \IEEEmembership{Fellow, IEEE},
    Wenpeng Zhu, \\
    Jing Huo \orcidlink{0000-0002-8504-455X}, \IEEEmembership{Member, IEEE}, 
    Yang Gao \orcidlink{0000-0002-2488-1813}, \IEEEmembership{Senior Member, IEEE},
    Depeng Wang,
    Haitao Wan,
    Xi Yang, \\
    Fanyu Meng \orcidlink{0009-0007-7120-7075},
    Yuyao Zhang \orcidlink{0009-0002-6637-4059},
    Ji Qi and
    Xiangyu Zhou
    \thanks{
    \textit{(Taihang Zhen and Jialiang Hong contributed equally to this work.) }
    }
    \thanks{
    \textit{(Corresponding authors: Junlan Feng, Wenpeng Zhu, Jing Huo, Yang Gao and Ji Qi.)}
    }
    \thanks{
    Taihang Zhen is with the School of Electronic Science and Engineering, Nanjing University, Nanjing 210023, China (e-mail: taihangzhen@smail.nju.edu.cn).
    }
    \thanks{
    Jialiang Hong, Wenpeng Zhu, Depeng Wang, Haitao Wan, Xi Yang, Ji Qi and Xiangyu Zhou are with the China Mobile (Suzhou) Software Technology Co., Ltd., Suzhou 215000, China (e-mail: \{hongjialiang, zhuwenpeng, wangdepeng, wanhaitao, yangxi, qiji, zhouxiangyu\}@cmss.chinamobile.com).
    }
    \thanks{
    Kai Chen, Guang Yang and Jing Huo are with the State Key Laboratory for Novel Software Technology, Nanjing University, Nanjing 210023, China (e-mail: \{kaichennju, yangg\}@smail.nju.edu.cn; huojing@nju.edu.cn).
    }
    \thanks{
    Junlan Feng, Fanyu Meng and Yuyao Zhang are with the JIUTIAN Research, China Mobile, Beijing 100000, China (e-mail: \{fengjunlanit, mengfanyuit01, zhangyuyaoit\}@chinamobile.com).
    }
    \thanks{
    Yang Gao is with the School of Intelligence Science and Technology, Nanjing University, Suzhou 215163, China (e-mail: gaoy@nju.edu.cn).
    }  
}

\markboth{Journal of \LaTeX\ Class Files,~Vol.~18, No.~9, September~2020}%
{How to Use the IEEEtran \LaTeX \ Templates}

\maketitle

\begin{abstract}
Large reasoning models (LRMs) often exhibit overthinking, producing verbose Chain-of-Thought (CoT) traces that increase inference cost and obscure the underlying reasoning process. Existing CoT compression methods mainly rely on global length rewards, which conflate necessary intermediate reasoning with redundant text and may therefore compromise reasoning fidelity. This paper revisits overthinking from a semantic-efficiency perspective and decomposes CoT redundancy into two distinct forms: internal redundancy, defined as informational stagnation before the first correct answer, and external redundancy, defined as superfluous continuation after the first correct answer. Based on this decomposition, we propose a dual-penalty reinforcement learning framework that separately optimizes reasoning progress and termination behavior. Specifically, a sliding-window semantic similarity metric penalizes low-progress reasoning segments, while a normalized external-redundancy metric discourages post-answer continuation.
Experiments on GSM8K, MATH500, and AIME24 across different model scales show that our method reduces average reasoning length by 41.3\% on the 1.5B model and 40.1\% on the 7B model, while preserving competitive accuracy and achieving the best overall accuracy-efficiency score among evaluated baselines. The learned compression behavior further transfers to out-of-domain reasoning tasks, including GPQA and LiveCodeBench. More importantly, our analysis reveals a clear asymmetry between the two redundancy types: external redundancy can be largely removed with little performance loss, whereas internal redundancy compression follows a sensitive accuracy-efficiency trade-off. These results suggest that effective CoT compression should optimize semantic efficiency rather than sequence length alone, offering a principled route toward more concise, efficient, and interpretable LRMs. Our code link is https://github.com/HenryZhen97/Reconsidering-Overthinking.
\end{abstract}

\begin{IEEEkeywords}
Large language models, large reasoning models, chain-of-thought reasoning, reinforcement learning, semantic redundancy, efficient reasoning.
\end{IEEEkeywords}

\section{Introduction}

Large reasoning models (LRMs) \cite{jaech2024openai,guo2025deepseek} have substantially advanced complex problem solving by generating explicit Chain-of-Thought (CoT) traces. These traces expose intermediate steps and enable models to navigate multi-step logical dependencies that are difficult to solve through direct answer generation. However, the same mechanism also introduces a growing reliability and efficiency challenge: LRMs often overthink, producing long reasoning traces that contain repeated derivations, low-information reformulations, or unnecessary verification after the answer has already been obtained \cite{chen2024not, sui2025stop}. Such verbosity increases inference cost, weakens the interpretability of CoT traces, and makes it harder to identify the reasoning steps that actually support the final answer.

\begin{figure*}
  \centering
  \includegraphics[width=0.9\linewidth]{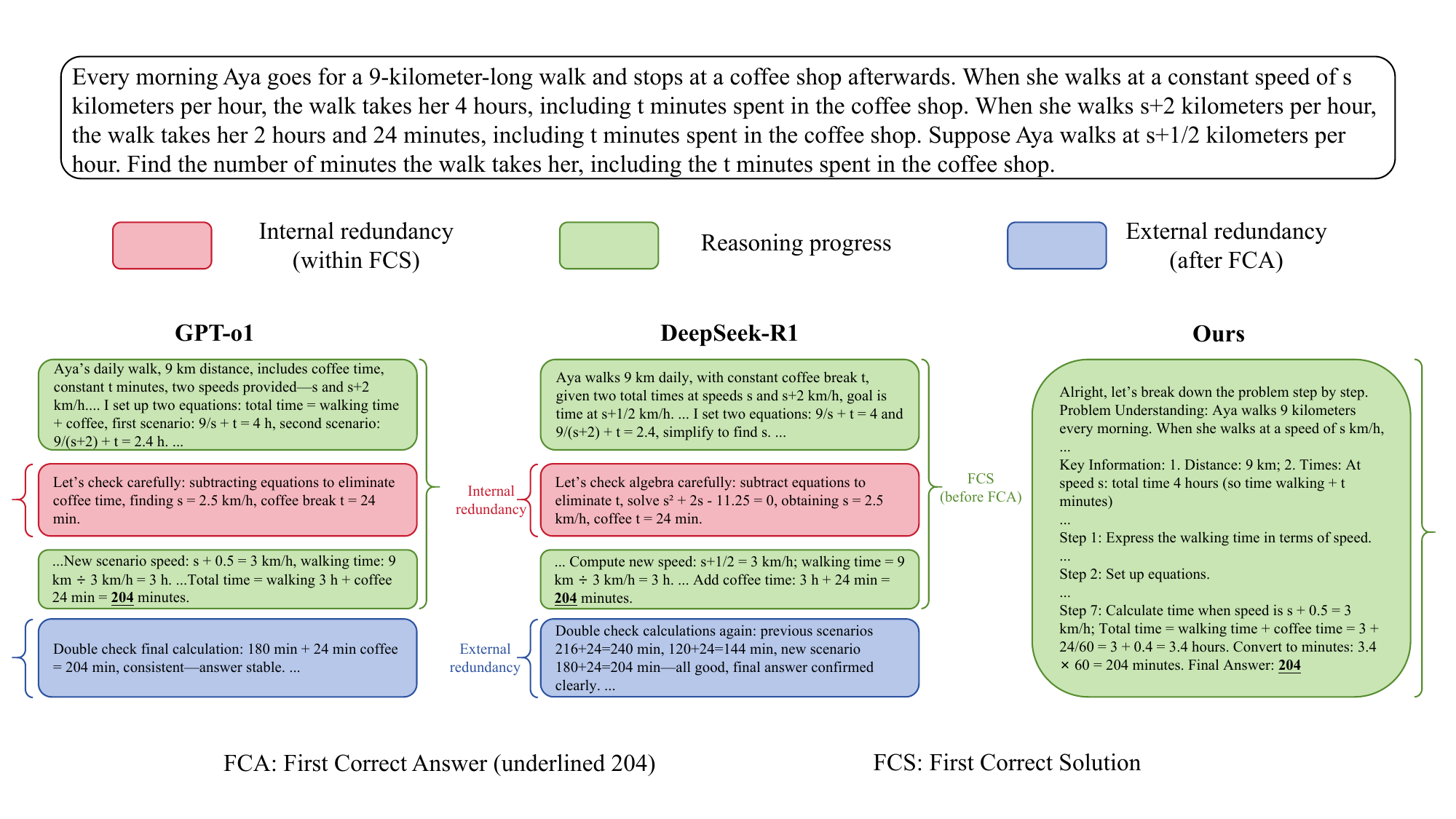}
  \caption{Illustration of CoT redundancy decomposition on an AIME24 example. The first occurrence of the correct final answer, marked by the underlined ``204'', is defined as the First Correct Answer (FCA). The prefix up to the FCA constitutes the First Correct Solution (FCS), where repeated or low-progress reasoning is regarded as internal redundancy. The continuation after the FCA is defined as external redundancy, reflecting unnecessary post-answer verification or re-derivation. Compared with GPT-o1 and DeepSeek-R1, our method yields a more compact FCS and effectively suppresses redundant continuation after the answer.}
  \label{Fig1}
\end{figure*}

Recent work has begun to address this problem through CoT compression, most commonly by adding reinforcement learning (RL) objectives that penalize total sequence length \cite{liu2025efficient, sheng2026reasoning, wang2025wait, wang2025thoughts}. Although length-based rewards are simple and often effective at reducing token counts, they treat reasoning as a one-dimensional budget allocation problem \cite{luo2025o1, arora2026training}. This abstraction overlooks a key fact: not all tokens play the same role in a reasoning trace. Some tokens encode indispensable intermediate steps, whereas others merely restate an existing state or continue after the final answer. Compressing both with the same global penalty can therefore remove useful reasoning along with redundancy, yielding a brittle trade-off between conciseness and accuracy.

We argue that overthinking should instead be analyzed through the semantic function of each segment in the reasoning process. To this end, we partition CoT redundancy according to the First Correct Answer (FCA), the earliest point at which the model states the correct final answer \cite{chen2024not}. This perspective separates two fundamentally different forms of inefficiency:
\begin{enumerate}
\item \textbf{Internal Redundancy}: informational stagnation within the First Correct Solution (FCS), where the model revisits semantically similar content, restates premises, or over-elaborates a transition without advancing the logical state.
\item \textbf{External Redundancy}: the post-answer tail, namely unnecessary continuation, re-derivation, or verification after the correct answer has already been reached.
\end{enumerate}

Representative examples in Figure \ref{Fig1} illustrate that both redundancy types can appear within the same response, yet they have different implications. Internal redundancy concerns reasoning efficiency: the model spends excessive tokens on limited logical progress. External redundancy concerns termination awareness: the model fails to stop after reaching the answer. This distinction is important because the two forms should not be penalized identically. This suggests that external redundancy is a safer compression target once the answer is reached, whereas excessive removal of internal redundancy may collapse intermediate reasoning steps that remain necessary for correctness.

\begin{figure*}
  \centering
  \includegraphics[width=0.9\textwidth]{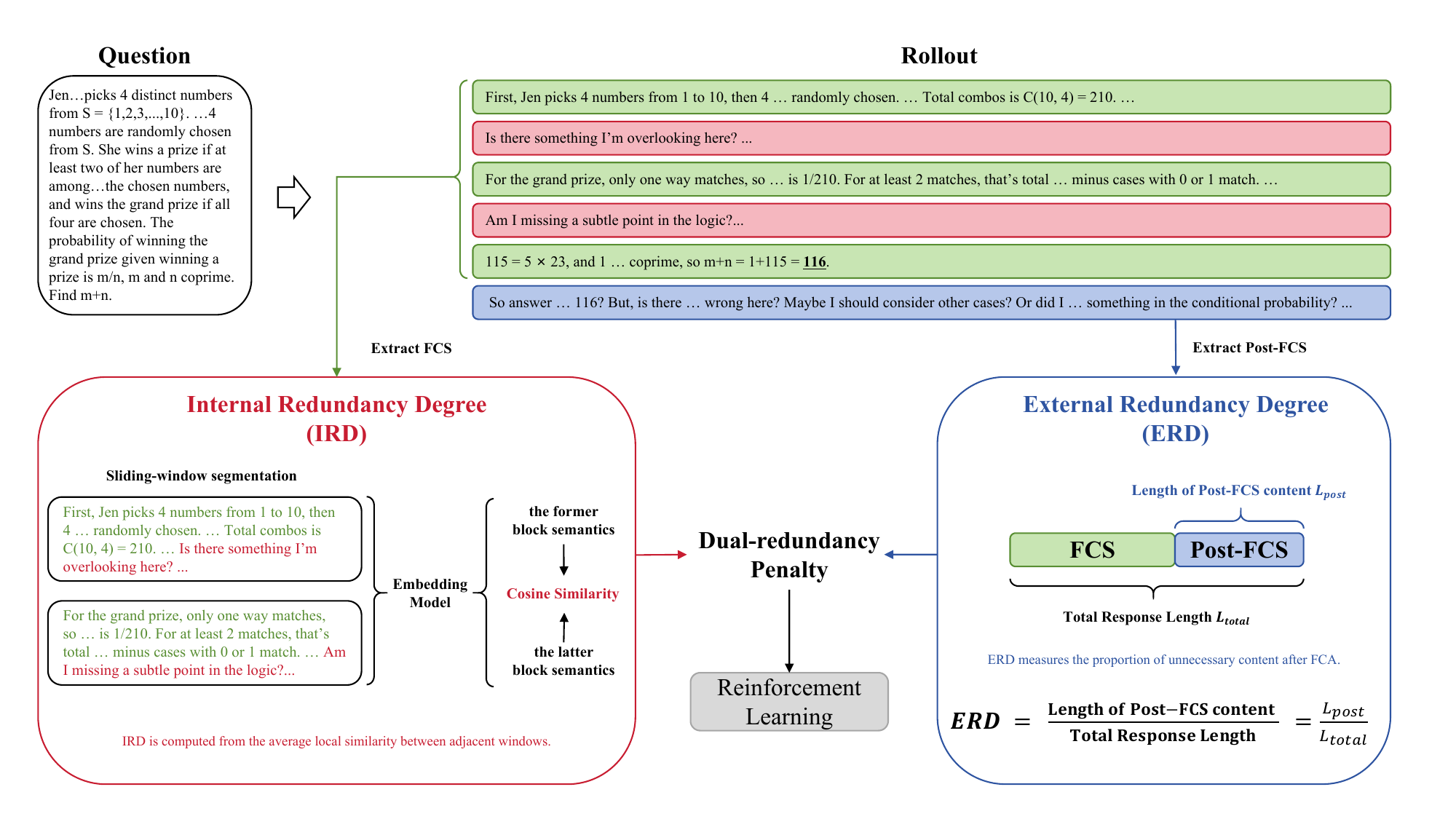}
  \caption{Overview of the proposed dual-redundancy penalty framework. Given a model rollout, the response is segmented at the First Correct Answer (FCA) into the First Correct Solution (FCS) and the post-FCS continuation. Internal Redundancy Degree (IRD) is estimated within the FCS by computing the average semantic similarity between adjacent sliding windows, capturing locally stagnant reasoning. External Redundancy Degree (ERD) measures the normalized proportion of unnecessary continuation after the FCA. The two redundancy signals are converted into complementary penalty terms and used to optimize the LLM through reinforcement learning.}
  \label{Fig2}
\end{figure*}

Based on this decomposition, we develop a \textbf{Dual-Redundancy Penalty} framework for RL-based CoT compression in Figure \ref{Fig2}. For internal redundancy, we compute a dynamic sliding-window semantic similarity score that identifies locally stagnant reasoning segments. The corresponding penalty uses an implicit threshold, discouraging low-progress segments while preserving a tolerance for semantically coherent reasoning. For external redundancy, we introduce a normalized proportion-based penalty that encourages the model to terminate after reaching the FCA. Compared with global length constraints, the proposed objective directly targets the semantic efficiency of the reasoning trajectory. It promotes concise reasoning without assuming that all tokens are equally dispensable.

We evaluate the proposed framework on GSM8K, MATH500, and AIME24 across two model scales. The results show that our method reduces average reasoning length by 41.3\% on the 1.5B model and 40.1\% on the 7B model, while maintaining competitive accuracy and achieving the best overall accuracy-efficiency score among the evaluated baselines. We further observe that the learned compression behavior transfers to out-of-domain reasoning tasks, including GPQA and LiveCodeBench. More importantly, controlled ablations reveal a clear asymmetry: \textbf{external redundancy can be largely eliminated with little performance loss, whereas internal redundancy compression follows a sensitive accuracy-efficiency frontier.} This finding supports the need for a decoupled treatment of CoT redundancy.

In summary, our contributions are:
\begin{itemize}
    \item We formulate CoT overthinking as two semantically distinct redundancy types, internal and external redundancy, shifting the compression objective from sequence-length minimization to semantic-efficiency optimization.
    \item We propose a dual-penalty RL framework that separately targets reasoning progress and termination behavior through a sliding-window internal redundancy penalty and a normalized external redundancy penalty.
    \item We demonstrate that the proposed method reduces average reasoning length by more than 40\% while preserving competitive accuracy, achieving stronger accuracy-efficiency trade-offs than global length-based baselines and transferring to out-of-domain reasoning tasks.
    \item We empirically identify an asymmetric effect of redundancy removal: external redundancy is comparatively safe to eliminate, whereas internal redundancy requires calibrated compression to preserve reasoning fidelity.
\end{itemize}

\section{Background}

\paragraph{Chain-of-Thought Reasoning and Overthinking} Chain-of-Thought (CoT) prompting \cite{wei2022chain} and related prompting strategies have become a central mechanism for eliciting step-by-step reasoning in large language models (LLMs), substantially improving performance on mathematical, symbolic, and commonsense reasoning tasks \cite{qiao2023reasoning}. More recent LRMs further scale this paradigm by using long deliberative traces to solve difficult problems \cite{jaech2024openai, guo2025deepseek}. However, longer reasoning does not always imply better reasoning. Prior studies have observed that o1-like models may overthink, producing unnecessarily long, repetitive, or self-verifying traces even for problems whose solutions have already been reached \cite{chen2024not, sui2025stop}. This line of work motivates the need to improve reasoning efficiency without undermining the intermediate steps that support correctness.

\paragraph{Efficient Reasoning and Test-Time Computation} A growing body of work studies how to allocate reasoning computation more adaptively. Some methods adjust reasoning depth according to task difficulty or planning signals \cite{sheng2026reasoning, shen2025dast, yang2025think}, while others prune or skip unnecessary thinking tokens during inference or training \cite{wang2025wait, han2025token, liu2024can, ma2025cot}. These studies show that reasoning traces contain compressible components \cite{donge2023data}, but they often focus on when or how much to think rather than on what kind of redundancy is being removed. As a result, the semantic roles of different redundant segments remain underexplored.

\paragraph{RL-Based CoT Compression} Reinforcement learning has recently become a prominent tool for training models to reason more concisely. Representative methods penalize total response length, impose token budgets, or shape rewards according to length-efficiency objectives \cite{team2025kimi, arora2026training, luo2025o1, hou2025thinkprune, cheng2025optimizing, liu2026learn}. These approaches are effective at reducing token counts \cite{lin2023survey}, but most of them treat redundancy as a single global property of the sequence. In contrast, our work decomposes CoT redundancy into internal redundancy before the first correct answer and external redundancy after it. This decomposition allows the reward to target reasoning progress and termination behavior separately and, more importantly, exposes an asymmetric accuracy-efficiency trade-off that is obscured by global length-based objectives.

\section{Redundancy Detection}
Redundant reasoning in LRMs can arise at different stages of a CoT trajectory \cite{han2025token, liu2024can, ma2025cot}. We use the First Correct Answer (FCA) as an operational boundary: tokens before the FCA form the First Correct Solution (FCS), while tokens after it constitute post-answer continuation. This segmentation separates redundancy that affects the reasoning process itself from redundancy that reflects a failure to terminate, enabling the two phenomena to be measured independently.

\subsection{External Redundancy}

\begin{figure}
\centering
\includegraphics[width=0.9\columnwidth]{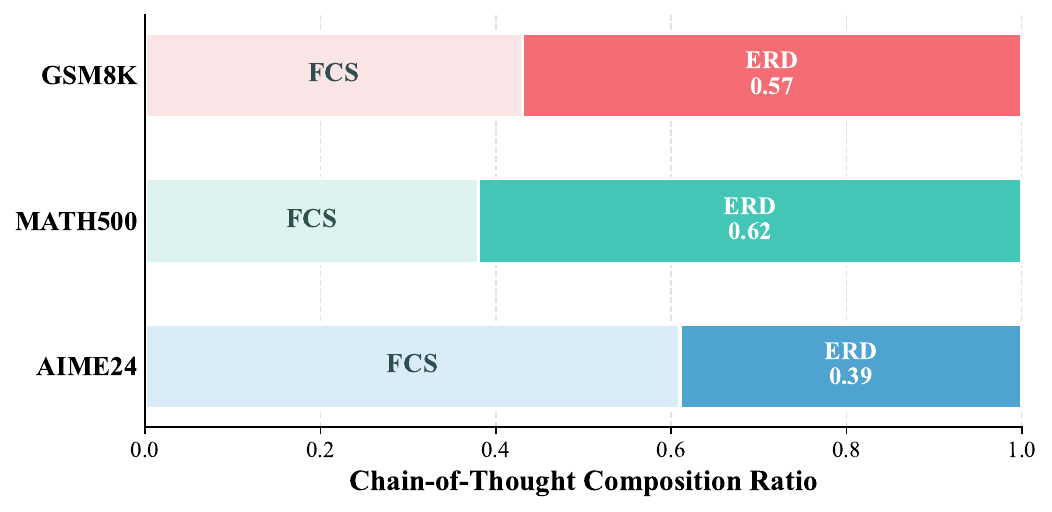}
\caption{External redundancy analysis on QwQ across math reasoning. Each bar decomposes a generated CoT response into the First Correct Solution (FCS) and the post-FCS continuation. A larger ERD indicates that a greater proportion of the response is spent on unnecessary reasoning after the correct answer has already appeared. ERD is reported for model-generated traces because human reference solutions are treated as having no post-answer continuation.}
\label{Fig3}
\end{figure}

Given a question $Q$ and a correct answer $A$, we define the FCA as the first sentence in a generated response that contains $A$ in the expected answer form. In our experiments, we locate this sentence by extracting candidate answer spans from the response, normalizing common mathematical answer formats such as boxed expressions and final-answer markers, and matching them against the reference answer. External redundancy is then defined as all content generated after the FCA sentence. This definition deliberately excludes trial-and-error reasoning before the FCA. Even if such reasoning is lengthy or inefficient, it still occurs before the model has produced the correct answer and should therefore be analyzed as part of the reasoning process rather than as post-answer continuation.

We quantify external redundancy with the External Redundancy Degree (ERD). ERD measures the fraction of a response that appears after the FCA rather than the absolute number of redundant tokens, reducing bias against problems that naturally require longer derivations. For a generated solution:
\begin{equation}
    \mathrm{ERD} = 1 - \frac{T_{fcs}}{T_{total}},
\end{equation}
where $T_{fcs}$ denotes the number of tokens up to and including the FCA sentence, and $T_{total}$ denotes the total number of tokens in the response. A higher ERD indicates that a larger fraction of the output is spent on post-answer continuation. Figure \ref{Fig3} shows that LRMs can retain substantial external redundancy even when the correct answer has already appeared.

\subsection{Internal Redundancy}

Internal redundancy is more subtle because it appears before the FCA, where the model is still constructing the solution. For a solvable question $Q$, assume an underlying logical trajectory $L = (Q, l_1, l_2, \dots, l_k, A)$, where each $l_i$ denotes a meaningful transition toward the answer. A generated solution $X$ can be viewed as a natural-language realization of this trajectory. Internal redundancy arises when the realization repeatedly elaborates the same logical state or transition without adding information that advances the trajectory.

Formally, given a logical trajectory $L$, we define internal redundancy as linguistic expansion within the FCS that does not facilitate progress toward a subsequent logical step. This includes repeated restatement, self-confirmation, and locally circular explanation. Unlike external redundancy, internal redundancy cannot be detected by answer position alone; it requires measuring whether adjacent reasoning segments provide new semantic content.

We operationalize this intuition through the following hypothesis:

\textit{For two solutions $X_a$ and $X_b$ that follow the same underlying trajectory $L$, if $X_a$ contains more internal redundancy than $X_b$, then adjacent local segments in $X_a$ will exhibit higher semantic similarity on average.}

Based on this hypothesis, we define the Internal Redundancy Degree (IRD) using dynamic sliding-window semantic similarity. The metric is designed to capture local semantic stagnation rather than global topic similarity.

Given an FCS $X$ partitioned into $N$ sentences $\{s_1, \dots, s_N\}$, we define a window size $w = \lfloor \alpha N \rfloor$ and a stride $t = \lfloor \beta N \rfloor$ with $\alpha>\beta$. For each window $W_i$, we compute its embedding $v_i = f_{\text{embed}}(W_i)$. IRD is the average cosine similarity between adjacent windows:
\begin{equation}
    \mathrm{IRD} = \frac{1}{M-1} \sum_{i=1}^{M-1} \cos(v_i, v_{i+1}),
\end{equation}
where $M$ is the number of windows. Higher IRD indicates stronger local similarity between neighboring segments, suggesting slower semantic progress within the FCS.

We use a proportional window because CoT responses vary substantially in length across problems. A fixed sentence window may be too coarse for short solutions and too local for long ones, whereas a ratio-based window preserves a comparable relative scale across responses. We empirically examine this design choice below.

\begin{figure*}
\centering
\includegraphics[width=0.85\linewidth]{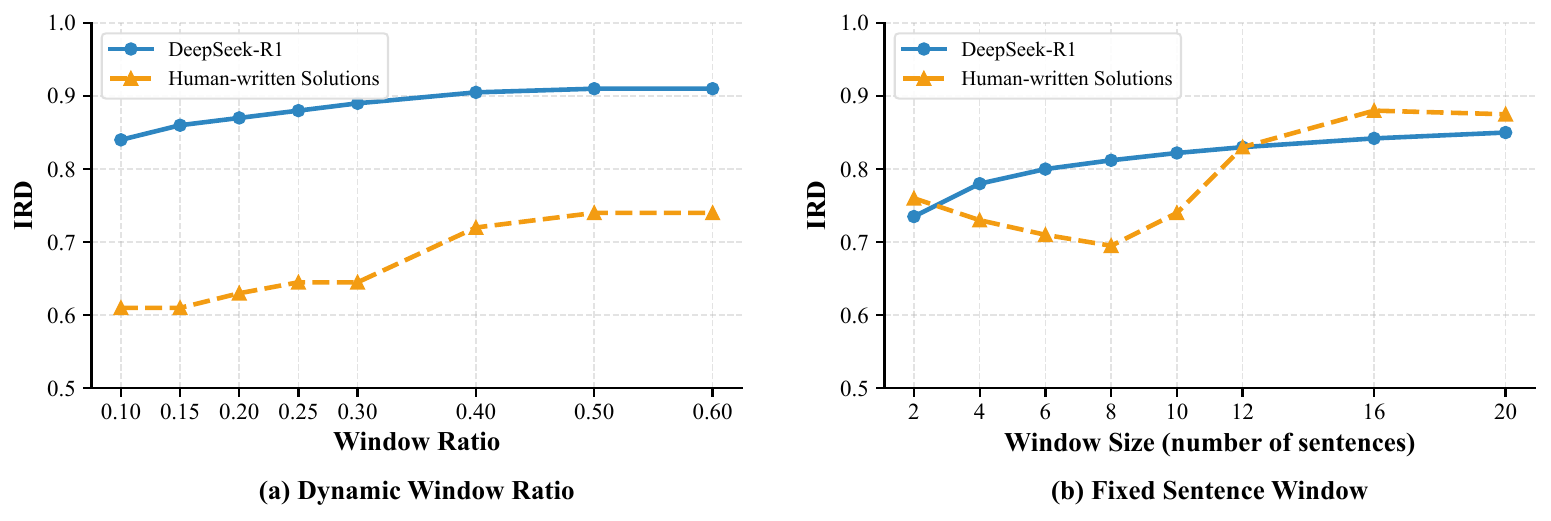}
\caption{Effect of window configuration on Internal Redundancy Degree (IRD) estimation. (a) Dynamic window ratio: the FCS is segmented using a proportional window size controlled by $\alpha$, with stride $\beta=\alpha/2$. As $\alpha$ increases, larger windows aggregate broader reasoning spans and weaken the separation between DeepSeek-R1 and human-written solutions. (b) Fixed sentence window: the FCS is segmented using a fixed number of sentences, which provides less stable separation due to variations in response length. These results motivate the use of a dynamic ratio-based sliding window for capturing local semantic stagnation.}
\label{Fig4}
\end{figure*}

\paragraph{Dynamic Window Size.} We first study the effect of the window-size ratio \(\alpha\) on local semantic similarity. We vary \(\alpha\) while setting the stride to half of the window size (\(\beta=\alpha/2\)), and compare DeepSeek-R1 solutions with human references on MATH500. As shown in Figure \ref{Fig4}a, the separation between DeepSeek-R1 and human-written solutions decreases as \(\alpha\) grows. Larger windows aggregate broader reasoning spans and can average out local stagnation signals.

\paragraph{Fixed Sentence Window.} We also evaluate an alternative formulation that uses a fixed number of sentences \(n\) as the window size, with stride \(n/2\). This design is simpler but ignores response-length variation across problems. Figure \ref{Fig4}b shows that fixed windows provide a weaker separation between DeepSeek-R1 and human references: the maximum margin is approximately \(0.1\), and the relative ordering is less stable across window sizes.

IRD should be interpreted as a relative measure of local semantic progress, not as an objective to be minimized to zero. Some similarity between adjacent windows is necessary for coherent reasoning, because consecutive steps must maintain context. Excessively low similarity may even indicate abrupt reasoning jumps. Therefore, IRD is most useful for comparing reasoning density across models, datasets, and training stages, and for defining a soft penalty that discourages stagnation while preserving coherent transitions.

\begin{figure*}
\centering
\begin{minipage}[b]{0.32\textwidth}
    \centering
    \includegraphics[width=\linewidth]{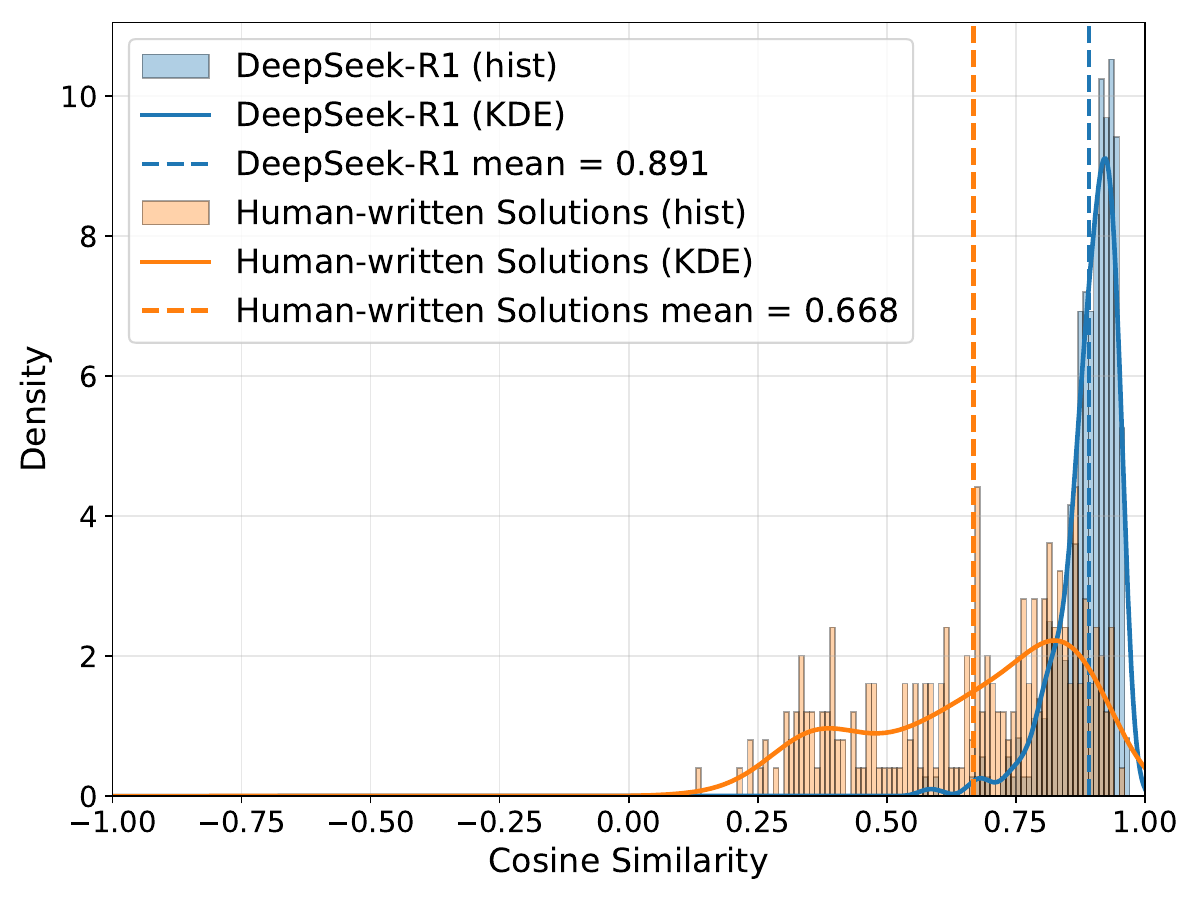}
\end{minipage}
\hfill
\begin{minipage}[b]{0.32\textwidth}
    \centering
    \includegraphics[width=\linewidth]{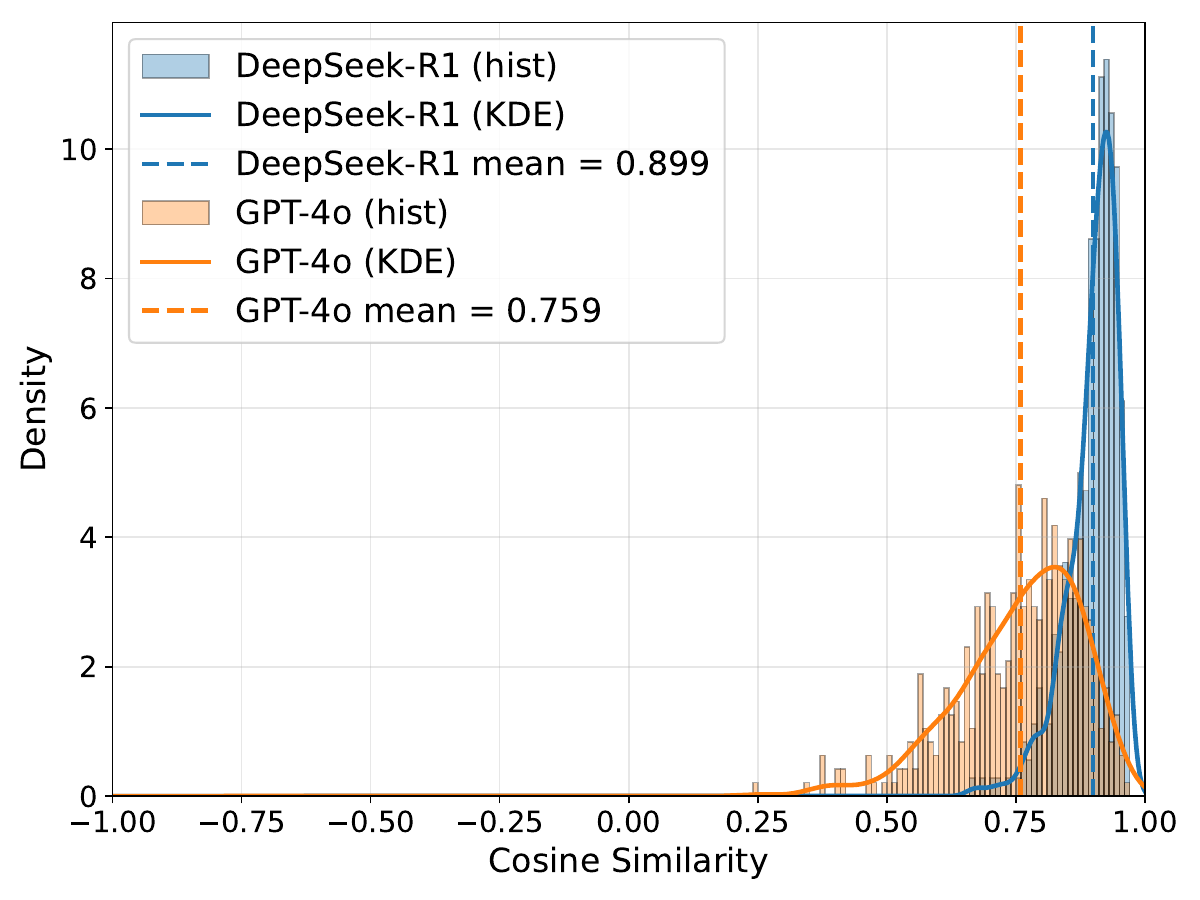}
\end{minipage}
\hfill
\begin{minipage}[b]{0.32\textwidth}
    \centering
    \includegraphics[width=\linewidth]{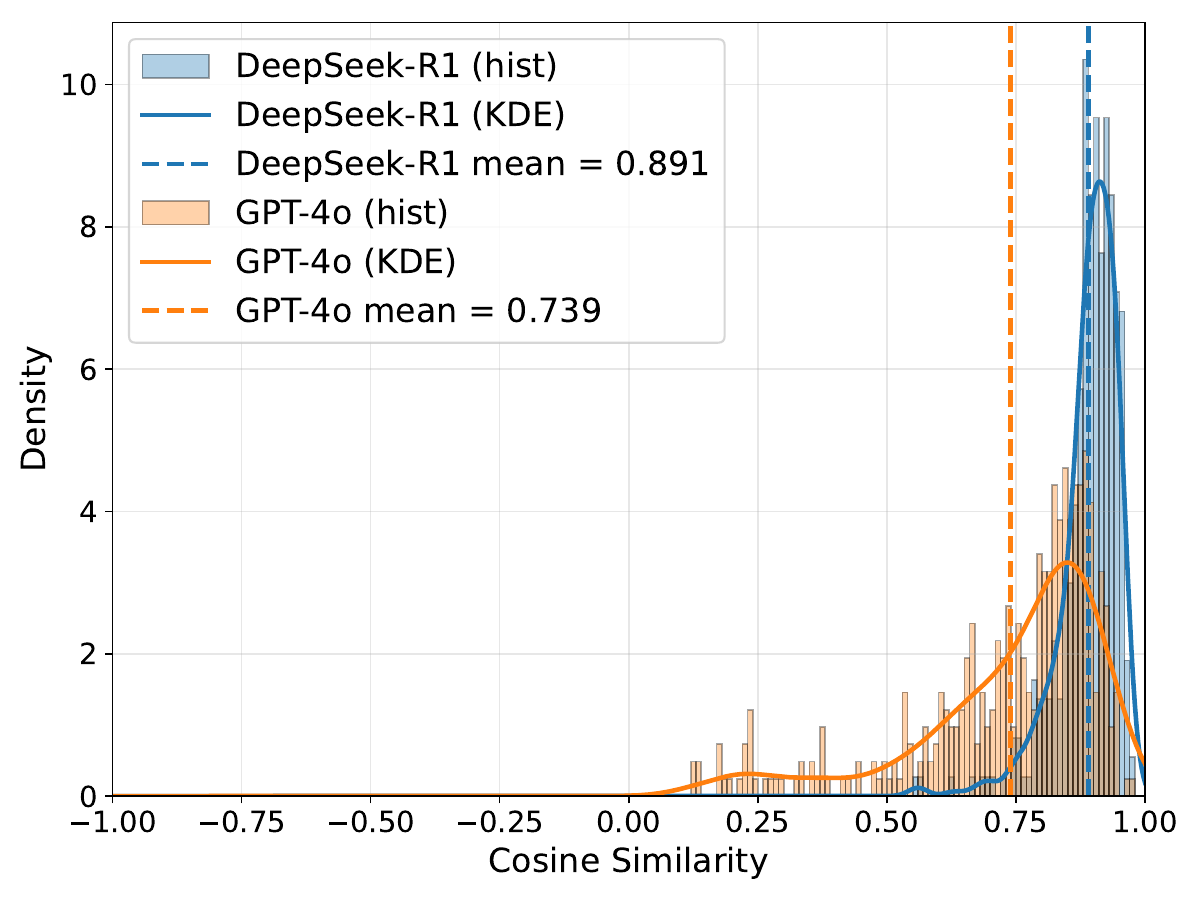}
\end{minipage}
\caption{Distribution of local semantic similarity for IRD analysis across AIME2024, LiveCodeBench, and GPQA. Each plot reports the cosine similarity between adjacent sliding windows within the First Correct Solution (FCS), with both histogram and kernel density estimation (KDE) curves. Compared with human-written solutions and GPT-4o outputs, DeepSeek-R1 shows distributions more concentrated in the high-similarity region and consistently higher mean similarity values, indicating stronger local semantic stagnation. These results support the use of IRD as a proxy for detecting internal redundancy in CoT reasoning.}
\label{Fig5}
\end{figure*}

To examine whether IRD captures meaningful differences in reasoning density, we compare DeepSeek-R1 traces with concise reference solutions, including human-written solutions and GPT-4o outputs. We randomly sample 20 instances from each of AIME24, GPQA \cite{rein2023gpqa}, and LiveCodeBench \cite{jain2025livecodebench}. As shown in Figure \ref{Fig5}, DeepSeek-R1 produces distributions concentrated in a higher-similarity region, whereas human references and GPT-4o outputs are more dispersed and generally lower. The mean IRD of DeepSeek-R1 is higher across all evaluated domains. These results support the hypothesis that redundant reasoning tends to manifest as elevated local semantic similarity, and justify using IRD as a training-time signal for internal redundancy.

\section{Dual-Redundancy Penalty}
The redundancy measures above provide training signals for improving CoT efficiency. We incorporate them into the RL reward as multiplicative reward modifiers rather than as standalone objectives. This design preserves answer correctness as the primary optimization target: redundancy penalties are applied only when the generated final answer is correct, so the model is not rewarded for producing short but incorrect reasoning traces.

Let \(R_{\mathrm{acc}}\) denote the accuracy reward. For a response whose final answer is correct, the total reward is
\begin{equation}
R_{\mathrm{total}} = R_{\mathrm{acc}} \cdot p_{\mathrm{int}} \cdot p_{\mathrm{ext}},
\end{equation}
where \(p_{\mathrm{int}}\in[0,1]\) and \(p_{\mathrm{ext}}\in(0,1]\) are reward multipliers derived from IRD and ERD, respectively. If the final answer is incorrect, we set \(R_{\mathrm{total}}=0\). This formulation encourages the model to improve reasoning efficiency only within the space of correct solutions.

\paragraph{Internal Redundancy Penalty} Internal redundancy is delicate because some local semantic continuity is necessary for coherent reasoning. Penalizing IRD linearly would pressure the model to reduce similarity even when adjacent steps are legitimately connected, increasing the risk of abrupt reasoning jumps. We therefore use a sharpened sigmoid to impose a soft threshold: responses below the threshold receive little penalty, whereas highly stagnant trajectories are penalized rapidly:
\begin{equation}
p_{\mathrm{int}} = 1 - \sigma(\mathrm{IRD}),
\end{equation}
\begin{equation}
\sigma(x) = \frac{1}{1 + e^{-k(x - c)}},
\end{equation}
where \(k\) controls the steepness and \(c\) controls the center of the transition. We set \(k=20\) and \(c=0.7\) in all experiments. With this configuration, \(p_{\mathrm{int}}\) remains close to one when IRD is in a moderate range, but decreases sharply when local semantic similarity becomes excessive. The multiplier therefore acts as an implicit tolerance mechanism: it discourages stagnant reasoning while avoiding direct pressure to eliminate all local similarity.

\paragraph{External Redundancy Penalty}  
External redundancy is less ambiguous because it occurs after the correct answer has already appeared. We therefore use a direct normalized multiplier based on ERD:  
\begin{equation}
p_{\mathrm{ext}} =  1 - \mathrm{ERD},
\end{equation}
which is equivalent to \(1 - T_{fcs}/T_{total}\). This multiplier equals one when the response terminates at the FCA and decreases as the post-answer tail grows. Because it depends on the proportion of post-answer content rather than absolute length, it discourages unnecessary continuation without penalizing problems that require longer pre-answer derivations.

Together, \(p_{\mathrm{int}}\) and \(p_{\mathrm{ext}}\) target complementary failure modes. The internal multiplier improves the density of reasoning before the answer, while the external multiplier improves termination behavior after the answer. This decoupling is essential: a global length reward cannot distinguish a useful intermediate step from a redundant post-answer continuation, whereas the proposed reward assigns penalties according to the semantic role and position of each segment.

\section{Experiment}
\subsection{Training Setup}
We implement training with verl \cite{sheng2025hybridflow} and optimize models using Group Relative Policy Optimization (GRPO) \cite{shao2024deepseekmath}. Experiments are conducted on 64 NVIDIA A800 GPUs. The maximum response length is 16k tokens, the sampling temperature is 0.6, and top-\(p\) is set to 1.0. For GRPO, we sample 8 responses per prompt and use a global batch size of 128.

We fine-tune DeepSeek-R1-Distill-Qwen-1.5B and DeepSeek-R1-Distill-Qwen-7B on the DeepScaleR dataset \cite{luo2025deepscaler}. Since our redundancy decomposition requires locating the FCA, we retain training problems with numeric answers containing at least two digits. This filtering reduces ambiguity in answer extraction and improves the reliability of FCS/post-answer segmentation. In our implementation, \(R_{\mathrm{acc}}\) is a binary correctness reward.

\subsection{Baselines}
We compare our method with representative RL-based CoT compression baselines. These methods reduce reasoning length through global budgets, length rewards, or teacher-guided compression, and therefore provide a direct comparison to our redundancy-aware reward.
\begin{itemize}
    \item ThinkPrune \cite{hou2025thinkprune} implements a hard token-budget truncation during RL, compelling the model to condense reasoning within a fixed length.
    \item LC-R1 \cite{cheng2025optimizing} utilizes an auxiliary LLM to provide external supervision, rewarding the model based on the compression ratio between the original and teacher-distilled responses.
    \item Laser-DE \cite{liu2026learn} employs a soft-margin reward mechanism, incentivizing correct outputs that fall below a predefined target length within a large context window.
    \item Training \cite{arora2026training} exploits intra-sample competition during RL, assigning higher rewards to shorter trajectories among multiple correct completions.
\end{itemize}

\subsection{Evaluation Setup}
We evaluate on three mathematical reasoning benchmarks of increasing difficulty: GSM8K \cite{cobbe2021training}, MATH500 \cite{hendrycks2021measuring}, and AIME24. During inference, the maximum generation length is 16k tokens. We use temperature 0.6. For GSM8K and MATH500, we sample \(n=4\) responses per problem; for AIME24, we sample \(n=64\) responses because the benchmark contains substantially fewer problems.

For token statistics, we exclude the terminal answer statement and count only the reasoning trace. This convention avoids confounding reasoning efficiency with answer-formatting differences, especially for methods that produce shorter or truncated final conclusions.

To evaluate the trade-off between reasoning performance and computational cost, we adopt the Accuracy-Efficiency Score (AES) \cite{luo2025o1}. AES combines relative length reduction with relative accuracy change:
\begin{equation}
\text{AES} = 
\begin{cases} 
\alpha \cdot \Delta \text{Len} + \beta \cdot |\Delta \text{Acc}|, & \text{if } \Delta \text{Acc} \geq 0 \\
\alpha \cdot \Delta \text{Len} - \gamma \cdot |\Delta \text{Acc}|, & \text{if } \Delta \text{Acc} < 0
\end{cases}
\end{equation}
where the relative changes in length (\(\Delta \text{Len}\)) and accuracy (\(\Delta \text{Acc}\)) are calculated relative to the baseline:
\begin{equation}
    \Delta \text{Len} = \frac{\text{Len}_{\text{base}} - \text{Len}_{\text{model}}}{\text{Len}_\text{base}},
\end{equation}

\begin{equation}
\Delta \text{Acc} = \frac{\text{Acc}_\text{model} - \text{Acc}_\text{base}}{\text{Acc}_\text{base}}.
\end{equation}
Following the original implementation, we set \(\alpha=1\), \(\beta=3\), and \(\gamma=5\). Since \(\gamma>\beta>\alpha\), AES rewards length reduction but penalizes accuracy degradation more strongly, making it suitable for evaluating compression methods that must preserve reasoning fidelity.

\subsection{Main Results}

Table \ref{table1} reports the main results. Across the two model scales, our method achieves the best overall AES among the evaluated methods. On the 1.5B model, it reduces the average reasoning length from 4595 to 2698 tokens, a 41.3\% reduction, while improving average accuracy across the three mathematical benchmarks. On the 7B model, it reduces the average reasoning length from 3640 to 2181 tokens, a 40.1\% reduction, while maintaining competitive accuracy.

The gains are not explained by length reduction alone. Several baselines substantially shorten responses but retain high IRD or ERD, indicating that global length pressure can reduce token count without consistently improving reasoning density or termination behavior. For example, LC-R1 produces very short GSM8K traces for the 7B model, but its accuracy drops noticeably. In contrast, our method jointly reduces IRD and ERD while preserving stronger accuracy-efficiency trade-offs. This pattern supports the central claim that effective CoT compression requires targeting the semantic structure of redundancy rather than total length alone.

The improvement is especially visible on AIME24, where over-compression can easily harm accuracy. Our method obtains the best AIME24 AES for both model scales and improves AIME24 accuracy over the baseline on both the 1.5B and 7B models. These results suggest that redundancy-aware compression can reduce unnecessary reasoning without removing the intermediate steps needed for more difficult problems.

\begin{table*}
\centering
\caption{Performance comparison with representative CoT compression baselines on GSM8K, MATH500, and AIME24. We report accuracy (Acc), average reasoning length (Tokens), Internal Redundancy Degree (IRD), External Redundancy Degree (ERD), and Accuracy-Efficiency Score (AES) for two model scales. Tokens exclude the terminal answer statement, and IRD/ERD values are multiplied by 100 for  readability. The overall AES summarizes the accuracy-efficiency trade-off across all benchmarks.}
\resizebox{1\linewidth}{!}{
\begin{tabular}{lcccccccccccccccc}
     \textbf{Model} & \multicolumn{5}{c}{\textbf{GSM8K}} & \multicolumn{5}{c}{\textbf{MATH500}} & \multicolumn{5}{c}{\textbf{AIME24}} & \multicolumn{1}{c}{\textbf{Overall}}\\
    \hline
     &Acc&Tokens&IRD&ERD&AES&Acc&Tokens&IRD&ERD&AES&Acc&Tokens&IRD&ERD&AES&AES\\
    \hline
    \multicolumn{16}{c}{\textit{DeepSeek-R1-Distill-Qwen-1.5B}} \\
    \hline
    Baseline & 84.1 & 1555 & 73.7 & 43.0 & / & 82.2 & 3549 & 77.5 & 55.2 & / & 28.5 & 8681 & 71.4 & 28.6 & / & /\\  
    ThinkPrune-4k & 86.1 & 910 & 77.9 & 40.1 & 0.49 & 83.7 & 2101 & 73.2 & 39.8 & 0.46 & 28.6 & 6431 & 75.2 & 21.0 & 0.27 & 1.22 \\
    LC-R1 & 82.5 & 507 & 67.2 & 19.3 & 0.58 & 79.6 & 1673 & 75.8 & 22.5 & 0.37 & 24.2 & 5075 & 79.6 & 20.4 & -0.34 & 0.61 \\
    Laser-DE & 86.4 & 971 & 74.3 & 37.5 & 0.46 & 83.6 & 2282 & 78.0 & 36.3 & 0.41 & 32.7 & 7268 & 73.5 & 22.2 & \underline{0.60} & \underline{1.47} \\
    Training & 81.0 & 292 & 61.6 & 7.8 & \underline{0.63} & 82.8 & 1543 & 65.5 & 14.5 & \underline{0.59} & 28.5 & 7049 & 73.2 & 17.4 & 0.21 & 1.13 \\
    \textbf{Ours} & 84.9 & 513 & 49.6 & 5.7 & \textbf{0.70} & 83.8 & 1505 & 51.0 & 7.9 & \textbf{0.63} & 34.0 & 6077 & 72.5 & 10.9 & \textbf{0.88} & \textbf{2.21} \\    
    \hline
    \multicolumn{16}{c}{\textit{DeepSeek-R1-Distill-Qwen-7B}} \\
    \hline
    Baseline & 91.1 & 844 & 70.0 & 36.0 & / & 91.2 & 2836 & 78.1 & 51.6 & / & 52.3 & 7241 & 77.8 & 31.1 & / & / \\
    ThinkPrune-4k & 92.8 & 716 & 70.5 & 36.0 & 0.21 & 89.7 & 1683 & 77.9 & 36.1 & 0.32 & 50.4 & 5723 & 79.2 & 14.6 & 0.03 & 0.56 \\
    LC-R1 & 87.5 & 152 & 61.8 & 4.9 & \textbf{0.62} & 87.5 & 1201 & 65.8 & 7.0 & 0.37 & 52.7 & 6087 & 79.1 & 10.2 & 0.18 & 1.17 \\
    Laser-DE & 93.3 & 637 & 68.2 & 31.1 & 0.32 & 92.1 & 1402 & 77.0 & 30.1 & \textbf{0.54} & 52.7 & 5061 & 80.5 & 11.8 & \underline{0.32} & \underline{1.18} \\
    Training & 91.2 & 387 & 65.1 & 14.6 & 0.54 & 91.0 & 2090 & 76.3 & 38.3 & 0.25 & 50.8 & 6669 & 78.8 & 23.1 & -0.06 & 0.73 \\
    \textbf{Ours} & 90.9 & 318 & 51.8 & 6.5 & \underline{0.61} & 89.8 & 1200 & 58.7 & 6.1 & \underline{0.50} & 53.2 & 5025 & 77.4 & 3.7 & \textbf{0.36} & \textbf{1.47} \\
    \hline
\end{tabular}
}
\label{table1}
\end{table*}

\begin{figure*}
\centering
\includegraphics[width=0.85\linewidth]{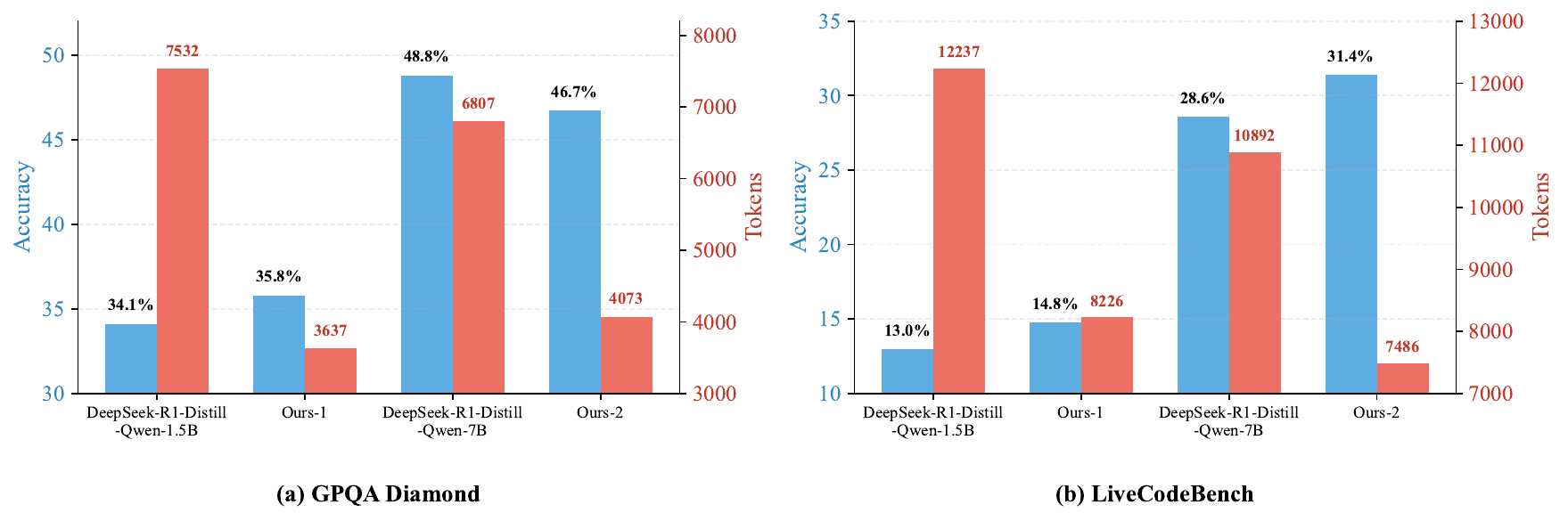}
\caption{Out-of-domain generalization on GPQA Diamond and LiveCodeBench. We compare the baseline DeepSeek-R1-Distill-Qwen models with the corresponding models trained using our dual-redundancy penalty at both 1.5B and 7B scales. For each benchmark, accuracy and average reasoning tokens are reported to evaluate whether the compression behavior learned from mathematical reasoning transfers to scientific question answering and code reasoning. Our method substantially reduces reasoning length while maintaining competitive out-of-domain task performance.}
\label{Fig6}
\end{figure*}

\subsection{Cross-Domain Generalization}

To examine whether the learned compression behavior transfers beyond the training distribution, we evaluate the trained models on two out-of-domain (OOD) benchmarks: GPQA Diamond \cite{rein2023gpqa}, which emphasizes graduate-level scientific reasoning, and LiveCodeBench \cite{jain2025livecodebench}, which evaluates code generation and programmatic reasoning.

As shown in Figure \ref{Fig6}, our method continues to shorten CoT traces on both OOD benchmarks while maintaining competitive task performance. Since the redundancy penalties are defined in terms of local semantic progress and post-answer continuation rather than domain-specific answer formats, the learned behavior is not tied to mathematical reasoning alone. These results suggest that the dual-penalty reward captures transferable properties of concise reasoning, although broader evaluation on more open-ended tasks remains an important direction for future work.

\section{Ablations}

\begin{figure*}
  \centering
  \includegraphics[width=1\linewidth]{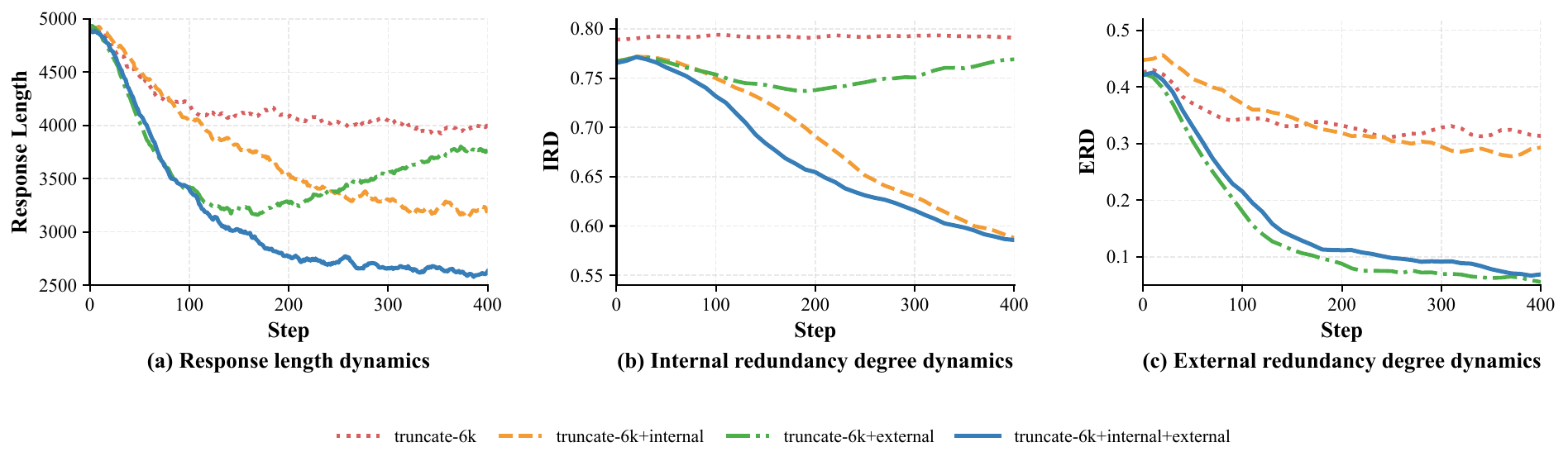}
  \caption{Ablation of internal and external redundancy penalties during training. We compare four configurations: length-based truncation, truncation with the internal redundancy penalty, truncation with the external redundancy penalty, and the full dual-penalty objective. (a) Response length dynamics. (b) Internal Redundancy Degree (IRD) dynamics. (c) External Redundancy Degree (ERD) dynamics. The internal penalty primarily reduces IRD, whereas the external penalty mainly reduces ERD. Combining both penalties achieves stronger compression while simultaneously reducing pre-answer semantic stagnation and post-answer continuation.}
  \label{Fig7}
\end{figure*}

\subsection{Internal vs. External Redundancy}
We first examine whether the two redundancy penalties affect the intended parts of the reasoning trace. Figure \ref{Fig7} compares four training configurations: a length-based truncation baseline, internal penalty only, external penalty only, and the full dual-penalty reward. Adding either redundancy penalty leads to shorter converged responses than the length-based baseline, while applying both penalties achieves the strongest compression in Figure \ref{Fig7}a . This indicates that the proposed reward improves efficiency beyond what can be obtained from a coarse length constraint alone.

Figures \ref{Fig7}b and \ref{Fig7}c further track IRD and ERD during training. The results show two consistent patterns:
\begin{enumerate}
    \item Limitations of global truncation: The truncation baseline reduces response length but does not consistently reduce either IRD or ERD, suggesting that length pressure alone does not reliably identify the source of redundancy.
    \item Partial decoupling of redundancy types: The internal penalty primarily reduces IRD, whereas the external penalty primarily reduces ERD. This behavior is consistent with the FCA-based segmentation: internal redundancy is located before the first correct answer, while external redundancy is located after it.
\end{enumerate}

\begin{figure*}
  \centering
  \includegraphics[width=0.75\linewidth]{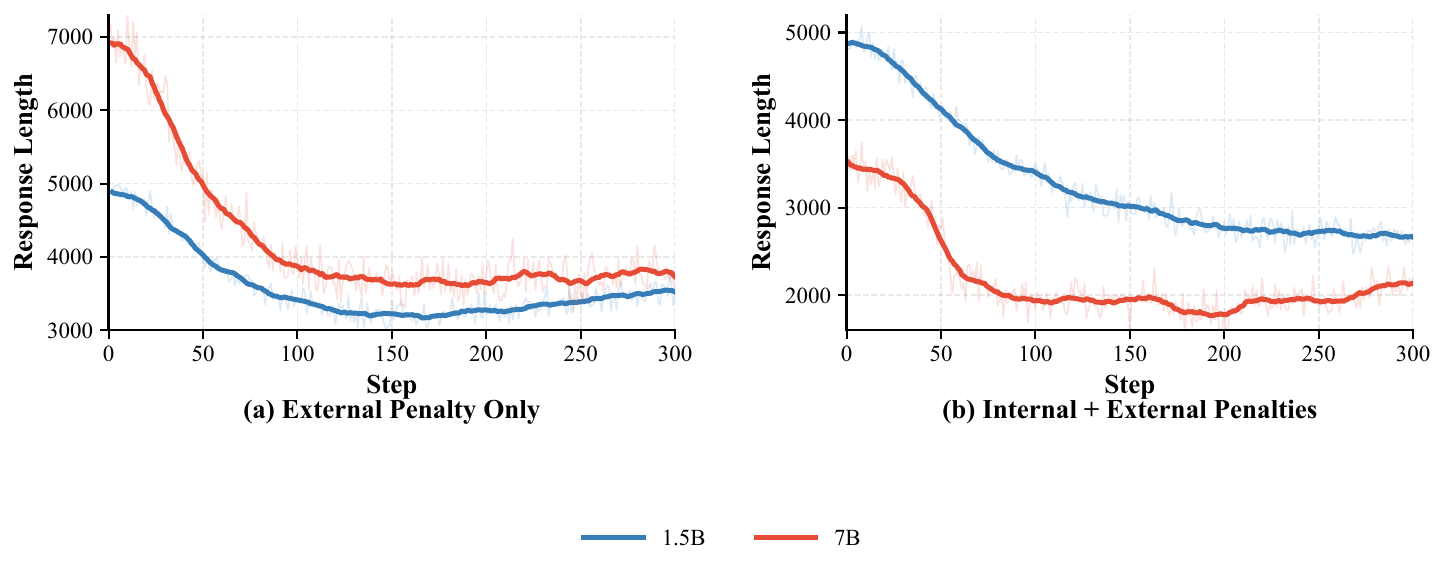}
  \caption{Response-length dynamics under different redundancy penalties at the 1.5B and 7B scales. (a) Using only the external redundancy penalty reduces response length in the early stage but may lead to a later length rebound, since the model can lower ERD by expanding the pre-answer reasoning portion. (b) Combining internal and external penalties suppresses this rebound by jointly discouraging pre-answer semantic stagnation and post-answer continuation.}
  \label{Fig8}
\end{figure*}

An additional effect appears when only the external penalty is applied: response length decreases early in training but later rebounds. Since ERD is a relative measure, the model can reduce the post-answer ratio not only by stopping earlier, but also by expanding the pre-answer portion. When the internal penalty is added, this rebound is suppressed, because unnecessary pre-answer expansion is also penalized. The 7B results in Figure \ref{Fig8} show the same qualitative behavior, supporting the complementarity of the two penalties.

\begin{figure*}
  \centering
  \includegraphics[width=1\linewidth]{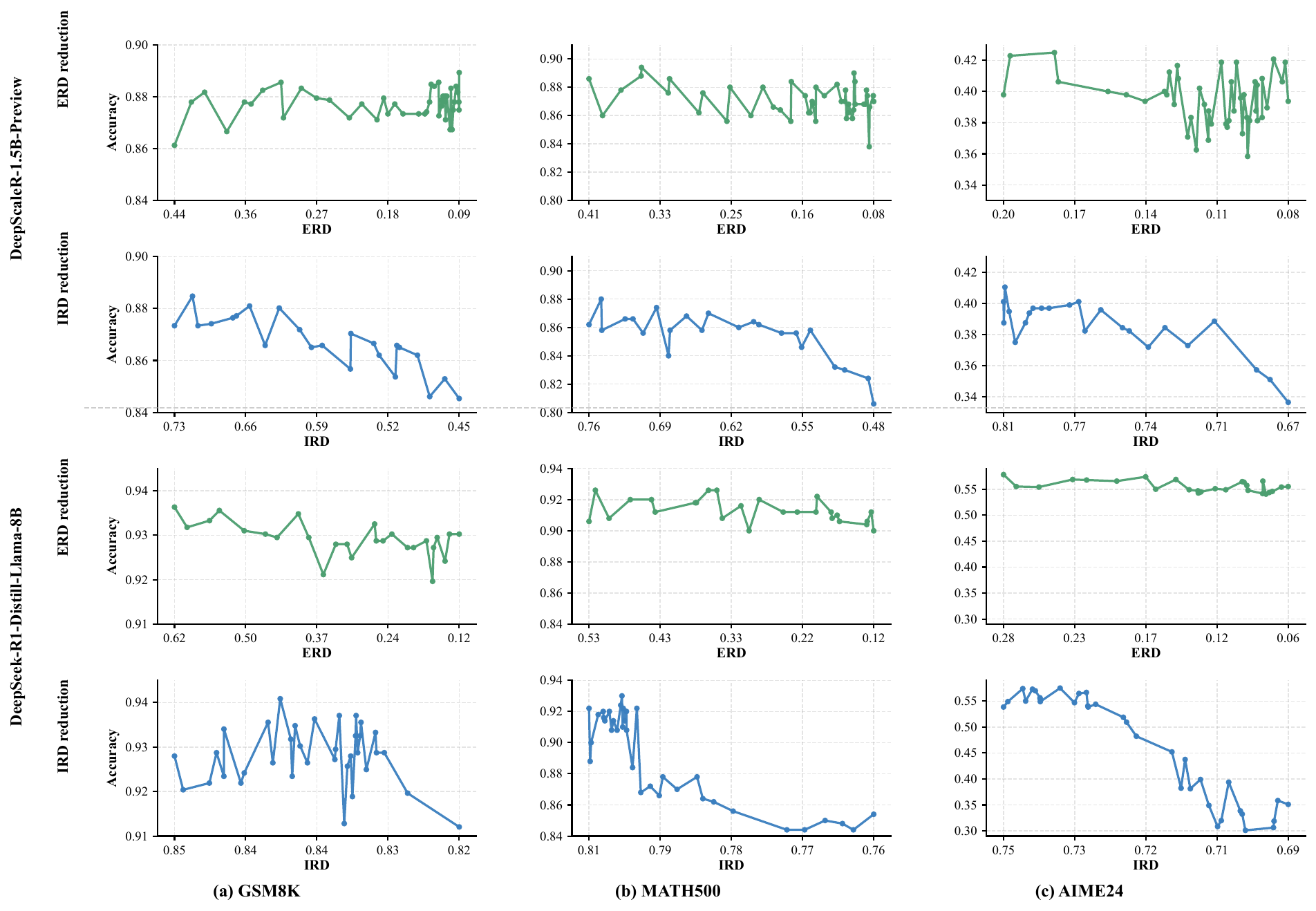}
  \caption{Effect of external and internal redundancy reduction on reasoning accuracy. We progressively reduce External Redundancy Degree (ERD) and Internal Redundancy Degree (IRD) on DeepScaleR-1.5B-Preview and DeepSeek-R1-Distill-Llama-8B, and report the resulting accuracy on GSM8K, MATH500, and AIME24. Reducing ERD has little impact on accuracy, suggesting that post-answer continuation is largely dispensable. In contrast, aggressive IRD reduction leads to noticeable accuracy degradation, especially on more challenging benchmarks such as AIME24, indicating that pre-answer compression must be carefully calibrated to preserve reasoning fidelity.}
  \label{Fig9}
\end{figure*}

\subsection{Analysis of Accuracy Drop} 
We next explore which redundancy type is responsible for the accuracy loss observed under aggressive compression. This question is important because shorter reasoning traces are useful only when the removed tokens are genuinely redundant. We conduct this analysis on DeepScaleR-1.5B-Preview. Since this model is already strong after RL scaling, it provides a relatively stable testbed for studying the effect of redundancy removal without conflating compression with large reward-driven accuracy gains.

\paragraph{Experimental Protocol} The ablation is conducted in two sequential stages.
\begin{itemize}
    \item Stage I (External): We apply only the external redundancy penalty with a 16k maximum response length. After 160 training steps, the average response length converges to approximately 4100 tokens, and ERD is substantially reduced.
    \item Stage II (Internal): Starting from the Stage I checkpoint, we add the internal redundancy penalty and continue training to progressively reduce IRD.
\end{itemize}

Figure \ref{Fig9} reveals a clear asymmetry between the two redundancy types. During Stage I, ERD decreases to 0.09 while accuracy remains nearly unchanged on GSM8K, MATH500, and AIME24. This indicates that post-answer continuation can be removed with little effect on correctness once the answer has already been reached. In contrast, Stage II shows that progressively reducing IRD leads to accuracy degradation, with the largest drop on AIME24. The result suggests that accuracy loss during CoT compression is mainly caused by over-compressing the pre-answer reasoning process rather than by removing the post-answer tail.

We attribute this behavior to the different semantic roles of the two redundancy types. External redundancy appears after the correct answer and therefore contributes little to deriving it. Internal redundancy, however, is interleaved with the reasoning trajectory. Aggressive IRD reduction may remove explanatory bridges between logical states, widening the semantic gap between adjacent steps and producing reasoning jumps that exceed the model's inference capacity. This interpretation is consistent with prior observations on thought leaps in CoT tuning \cite{xu2026mind}.

Similar trends are observed on DeepSeek-R1-Distill-Llama-8B in Figure \ref{Fig9}, suggesting that the asymmetric effect is not specific to a single model scale or backbone.

\section{Discussion}

The proposed decomposition is most directly applicable to tasks with objectively verifiable answers, such as mathematics. In these settings, the FCA provides a practical boundary between reasoning before the answer and continuation after the answer. For open-ended generation, however, there may be no single answer sentence that can serve as a reliable pivot. Extending the framework to such tasks would require task-dependent criteria for deciding when a response has satisfied the user intent and when further elaboration becomes redundant.

IRD should also be viewed as an operational proxy for internal redundancy rather than a complete model of logical progress. Sliding-window semantic similarity captures local stagnation and repeated reformulation, but it may miss higher-level reasoning cycles that use diverse wording. Conversely, some domains may require high local similarity to maintain formal precision or contextual continuity. The metric is also influenced by the embedding model used to represent reasoning segments. Future work could combine semantic similarity with symbolic state tracking, proof-step verification, or process-level supervision to better distinguish necessary elaboration from true logical redundancy.

Finally, our experiments focus on reasoning-intensive tasks where conciseness and correctness are the primary desiderata. In tasks that require exploration, planning under uncertainty, or creative generation, some apparent redundancy may help preserve alternatives or contextual nuance. The asymmetric finding in this work therefore should be interpreted within the scope of objective reasoning: post-answer continuation is usually expendable, whereas pre-answer compression must be calibrated carefully. This distinction provides a useful foundation for efficient reasoning, but broader task families may require adaptive definitions of redundancy.

\section{Conclusion}
We decompose CoT redundancy into internal redundancy before the first correct answer and external redundancy after it, and introduce a dual-penalty RL framework that separately targets reasoning progress and termination behavior. The central finding is that the two redundancy types have asymmetric effects on reasoning fidelity. External redundancy is a comparatively safe compression target once the answer has been reached, whereas aggressive internal redundancy reduction can remove useful intermediate structure and degrade accuracy. This suggests that efficient reasoning should not be optimized through sequence length alone, but through semantic criteria that respect the different roles of reasoning segments. We hope this perspective encourages future work on more reliable, interpretable, and efficient reasoning models.

\bibliographystyle{IEEEtran}
\bibliography{ref}

\end{document}